\pdfoutput=1

\documentclass[11pt]{article}

\usepackage[]{ACL2023}

\usepackage{times}
\usepackage{latexsym}
\usepackage{dblfloatfix}

\usepackage{booktabs}
\usepackage{adjustbox}
\usepackage[normalem]{ulem}
\useunder{\uline}{\ul}{}

\usepackage[T1]{fontenc}

\usepackage[utf8]{inputenc}

\usepackage{microtype}

\usepackage{inconsolata}

\title{Background Knowledge Grounding for Readable, Relevant, and Factual Biomedical Lay Summaries}

\author{Domenic Rosati \\
  scite.ai / Brooklyn, N.Y. \\
  Dalhousie University / Halifax N.S.}

\begin{document}
\maketitle
\begin{abstract}

Communication of scientific findings to the public is important for keeping non-experts informed of developments such as life-saving medical treatments. However, generating readable lay summaries from scientific documents is challenging, and currently, these summaries suffer from critical factual errors. One popular intervention for improving factuality is using additional external knowledge to provide factual grounding. However, it is unclear how these grounding sources should be retrieved, selected, or integrated, and how supplementary grounding documents might affect the readability or relevance of the generated summaries. We develop a simple method for selecting grounding sources and integrating them with source documents. We then use the BioLaySum summarization dataset to evaluate the effects of different grounding sources on summary quality. We found that grounding source documents improves the relevance and readability of lay summaries but does not improve factuality of lay summaries. This continues to be true in zero-shot summarization settings where we hypothesized that grounding might be even more important for factual lay summaries.
\end{abstract}

\section{Introduction}

\begin{figure}[!t]
    \centering
    \noindent\fbox{%
    \parbox{\linewidth}{%
    \small
        \textbf{Lay Summary:} [Messenger RNAs carry the instructions necessary to synthesize proteins that do work for the cell]\textsubscript{background} In this work , we surveyed mRNA ends from 10 , 000 genes in immune cells from genetically distinct human individuals.
\noindent\rule{\linewidth}{0.5pt}
\textbf{Abstracts:} Virtually all messenger RNAs (mRNAs) in eukaryotes are cleaved and polyadenylated at their 3 ends. \\
\textbf{UMLS:} RNA is unique among biological macromolecules in that it can encode genetic information. \\ 
\textbf{Wiki Simple:} Messenger RNA carries a genetic message from the DNA to the protein making machinery of the cell. \\
\textbf{Wikipedia}: An mRNA molecule is transcribed from the DNA sequence and is later translated into protein. 
    }%
}%
    \caption{Lay summaries use background knowledge often not in the original paper. Example of different types of background knowledge from grounding sources.}
    \label{fig:example}
\end{figure}

Automatic lay summarization of biomedical research is a promising approach to help inform non-experts of vital scientific and clinical discoveries. However, known issues with factuality in automatic summarization systems \citep{gabriel_go_2021,maynez_faithfulness_2020} are still a barrier that prevents their safe deployment. To improve factuality, some have suggested using grounding sources with retrieval augmentation \citep{shuster_retrieval_2021,lewis_retrieval-augmented_2021,thoppilan2022lamda}. This has been show to help maintain factuality in biomedical nlp tasks \citep{guo_cells_2022} without harming readability or relevance. Additionally, \citep{guo_cells_2022} suggests that retrieval augmentation is especially helpful for lay summarization because those summaries need to provide necessary background knowledge (see figure \ref{fig:example}) such as definitions which are not often found in the source text.

In this paper we wanted to understand (i) how we might develop a retrieval augmentation solution for lay summarization when using whole scientific papers for models with limited input context length and (ii) what is the effect of different grounding sources that contain different types of background knowledge on readability, relevancy, and factuality.

\textbf{Contributions:} We develop (i) \textit{a simple method for selecting and using grounding sources for lay summarization}. We assess this method using the BioLaySum \citep{goldsack_making_2022} biomedical paper lay summarization dataset and find that (ii) \textit{grounding has the largest effect on readability} where in the zero-shot summarization setting definitional background knowledge from Unified Medical Language System (UMLS) and simplified encyclopedic background knowledge from Wikipedia Simple provide better readability scores. Contrary to popular opinion, we found that (iii) \textit{grounding does not improve factuality}.

\section{Method}
\label{sec:method}

Summarizing scientific papers requires more input tokens than a large language model can typically support due to memory constraints. The average token count for articles in BioLaySum are 8,963 tokens for PLOS and 13,942 tokens for eLife with articles up to 45,563 tokens. Since we want to explore the effect of grounding articles with additional retrieved sources there is an even greater need for large token input support. Because of these factors, the base model we use in our experiments is the Longformer Encoder-Decoder (LED) \citep{beltagy_longformer_2020} which supports an input token length of 16,384 tokens (see Appendix \ref{app:models} for additional training and inference details).

Our method was designed to test the effect of different grounding sources on downstream summarization quality. In addition to definitional background knowledge from UMLS and encyclopedic background knowledge from Wikipedia which were used in \citet{guo_cells_2022}, we introduced two other retrieval sources, \textit{Wikipedia Simple} for access to encyclopedic background knowledge in simpler terms and \textit{Scientific Abstracts} for access to further contextual background knowledge that might emulate the additional supplementary knowledge an expert has when crafting a lay summary. In all, we tested the following four grounding sources (1) \textit{UMLS} named entity definitions (2) \textit{Scientific Abstracts} (from Crossref) (3) \textit{Wikipedia} (English) (4)  \textit{Wikipedia Simple} (English). See Appendix \ref{app:grounding_sources} for a full description of these grounding sources and how they were used.

Our retrieval augmentation consisted of two steps: (i) retrieving and (ii) re-ranking documents,
First we took each sentence in the leading 1,024 tokens of the article (roughly corresponding to the abstract) and searched them using BM25 on indexes constructed for each grounding source (except \textit{UMLS} which uses another method discussed). Indexes were constructed using Pyserini \citep{Lin_etal_SIGIR2021_Pyserini}. The top 1 most relevant passage is selected and then added to a pool of candidate passages. In the case of \textit{Scientific Abstracts}, we remove the abstract of the document we are enhancing if it was in the pool. In the case of \textit{UMLS}, we follow \citet{guo_cells_2022} by using the scispaCy entity linker \citep{neumann_scispacy_2019} over the first 1,024 tokens and provide definitions for the UMLS named entities as the pool of candidate passages. 

The above procedure results in too many results to fit within context length. In order to resolve this, we rank the pool of candidates passages against the first 1,024 tokens of the using a crossencoder \citep{reimers_sentence-bert_2019} (see Appendix \ref{app:models} for details). Finally we construct our inputs by selecting the first 8,192 tokens of the original document and the top relevant grounding passages up to 8,192 tokens. A \texttt{<|SEARCH|>} token is inserted between the original document and the grounding passage and global attention is placed on the \texttt{<|SEARCH|>} token in order to assist with attention over the grounding passages. 

We also supplement all grounding sources with a bibliographic reference string containing the title, authors, and year of the paper being summarised. This was motivated by seeing many ground truth summaries which cited the source document by first authors name (For example: "Parks et al. analyzed data on US deaths between 1980 and 2016" which is the first sample in the eLife training subset of BioLaySum).

\begin{table*}[ht!]
\centering
\begin{adjustbox}{center,max width=\linewidth}
\begin{tabular}{@{}lllllllll@{}}
\toprule
            & DCRS         & FKGL         & bartscore    & summac      & bertscore    & rouge1       & rouge2       & rougeL       \\ \midrule
LED         & 12.36 &  \underline{15.51} & \textbf{-2.21} & \textbf{21.37} &  \underline{86.22} & \underline{45.01} & \underline{15.44} & \underline{24.15} \\
All         & \underline{12.29} &  \textbf{15.42} & -2.28 & \underline{21.36} &  \textbf{86.26} & \textbf{45.31} & \textbf{15.75} & \textbf{24.47} \\ \midrule
Wikipedia   & \textbf{12.27} &  15.58 & -2.24 & 21.17 &  86.15 & 44.70 & 15.01 & 23.81   \\
Abstracts   & 12.30 &  15.62 & -2.29 & 21.30 &  86.18 & 44.67 & 15.29 & 24.04 \\
UMLS        & 12.32 &  15.61 & -2.23 & 21.22 &  86.14 & 44.67 & 14.94 & 23.80          \\
Wiki Simple & 12.31 &  15.58 & \underline{-2.22} & 21.24 &  86.14 & 44.51 & 14.87 & 23.74         \\ \bottomrule
\end{tabular}
\end{adjustbox}
\caption{\label{tab:grounding_comparison}
Readability, factuality, and relevancy scores for different grounding sources compared against LED baseline.  Lower is better for DCRS and FKGL.
}
\end{table*}

\begin{table*}[b!]
\centering
\begin{adjustbox}{center,max width=\linewidth}
\begin{tabular}{@{}lllllllll@{}}
\toprule
             & DCRS          & FKGL          & bartscore      & summac         & bertscore      & rouge1         & rouge2         & rougeL         \\ \midrule
All            & 11.31 & 14.04 & \underline{-3.35} & \textbf{20.30} &  \underline{85.78} & 39.09 & 11.26 & 21.50 \\
Without           & 11.23 & 14.15 & \textbf{-3.08} & \textbf{20.30} &  \textbf{86.02} & \textbf{40.66} & \textbf{11.98} & \textbf{21.97} \\
Abstracts      & 11.45 & 14.48 & -3.39 & \underline{20.26} &  85.72 & \underline{39.41} & \underline{11.42} & \underline{21.52} \\
UMLS           & \textbf{10.80} & \textbf{13.42} & -3.37 & 20.25 &  85.56 & 38.40 & 10.00 & 20.69 \\
Wikipedia      & 11.35 & 14.11 & -3.49 & \underline{20.26} &  85.49 & 39.05 & 10.48 & 20.61 \\
Wik Simple     & \underline{10.88} & \underline{13.43} & -3.75 & 20.21 &  84.96 & 36.40 &  9.03 & 19.20 \\
 \bottomrule
\end{tabular}
\end{adjustbox}
\caption{\label{tab:gpt}
Zero-shot summarization setting exploring readability, factuality, and relevancy scores for each grounding source using GPT 3.5 Turbo.
}
\end{table*}

\section{Experiments}

\subsection{BioLaySum}

We experiment with the method above using the BioLaySum lay summarization dataset which contains 29,119 training samples (24,773 from PLOS and 4,346 from eLife) and 1,617 validation samples (1,376 from PLOS and 241 from eLife). See Appendix \ref{app:biolaysum} for more details. This dataset is used to evaluate the ability for models to provide factual, readable, and relevant lay summaries of biomedical research articles from research papers in PLOS and eLife which are paired with human written lay summaries. All of the methods use a LED base model trained for 4 epochs evaluated every 5,000 steps, like \citep{goldsack_making_2022} the checkpoint with the best ROUGE-2 is selected (see Appendix  \ref{app:models} for more details).

Relevancy was measured based on BERTScore, Rouge1, Rouge2, and RougeL. Factuality was measured using a BARTScore \citep{yuan_bartscore_2021} trained on the BioLaySum dataset as well as an unsupervised metric SummaC \citep{laban_summac_2022}. Readability scores used Dale-Chall Readability Score (DCRS) and Flesch-Kincaid Grade Level (FKGL) measures.

Table \ref{tab:grounding_comparison} compares \textit{LED}, a baseline model that is only trained on the original documents to generate the output lay summaries, and \textit{All}, where the model is trained on the original document and supplemented with passages from the retrieval sources.

The results in table \ref{tab:grounding_comparison} show a few trends, first that \textit{All} improves relevancy and readability over the \textit{LED} setting. However, these gains are not very large which indicates that the grounding sources were not a very important signal for the model. Additionally, despite our hypothesis, factuality is not improved with grounding.

\subsection{Analysis of Grounding Methods}

To assess the impact of various retrieval sources on summarization quality, we trained a model for each retrieval corpus (table \ref{tab:grounding_comparison}). The most noticeable difference is that articles grounding in passages from \textit{Scientific Abstracts} and \textit{Wikipedia} have the highest relevancy scores. Possibly due to a similarity in the text distributions of these sources with reference lay summaries. Other grounding sources have different genre roles such as definitions in the case of \textit{UMLS} or non-expert reference literature in the case of \textit{Wikipedia Simple}. All methods generally improve readability as measured by DCRS with \textit{Wikipedia} having the largest effect.

\subsection{Zero-shot Summarization}

The lack of differences between grounding sources inspired us to consider an experiment where a model might be more likely to use grounding sources. We designed a zero-shot summarization experiment (with GPT 3.5) using the same method from section \ref{sec:method} but with 2,048 tokens selected for the original document and the rest from the original article (see Appendix \ref{app:gpt} for more details).

The results in table \ref{tab:gpt} show that grounding sources make much more of a difference for zero-shot summaries than trained summaries. The \textit{Without} input, which means without any grounding sources, has the highest relevancy and factuality scores indicating that in a zero-shot setting grounding sources tend to provide some distraction. Interestingly, \textit{UMLS} and \textit{Wikipedia Simple} encourage more readable summaries than other methods which is what we would expect to find since \textit{UMLS} provides definitions which are vital to assisting non-experts with engaging scientific findings and \textit{Wikipedia Simple} provides plain language encyclopedic knowledge designed to be readable. As we saw above, \textit{Scientific Abstracts} as a grounding source allows us to construct more relevant summaries, perhaps due to \textit{Scientific Abstracts} preserving the scientific language and context that is still vital for lay summaries.

\section{Discussion}

Grounding the original document in retrieval results helped most with readability, marginally with relevance and not at all with factuality. We believe this gives credence to the idea that grounding for lay summarization is primarily helpful for providing the model with background information that helps explain concepts, define terms, and otherwise situate the reader with the necessary information to be able to understand a scientific finding. We saw that the \textit{UMLS} and  \textit{Wikipedia Simple} sources provided best effect on improving readability in the zero-shot summarization setting which is intuitive in that they provide clear definitions and encyclopedic background knowledge in simple terms. \textit{Scientific Abstracts} had the best effect of all sources on relevancy which is also intuitive because related abstracts such as ones with similar findings might help construct a more robust summary. These results indicate that we should continue to investigate the role of background information on improving readability and relevance in lay summarization. In particular, our retrieval method is quite simple and more sophisticated methods of retrieval such as dense passage retrieval \citep{izacard2022unsupervised} could be used to enhance the relevancy of grounding documents. Additionally, future work should investigate methods that learn more strongly to take adavantage of grounding sources.

The lack of improvement in factuality is an important aspect that future work should investigate. One explanation is that grounding sources could introduce factual or relevancy errors if the retrieved documents are irrelevant or incorrect and they end up being used in the generated summary. However, factuality metrics only measure the summary against original document, this means that statements that cannot be grounded in the original document may be penalized. This is an issue in lay summarization where there can be statements that are factual and provide necessary background knowledge but cannot be found in the original document. Future work should investigate methods of measuring factuality that are able to incorporate the necessary background knowledge when measuring the factuality of a lay summary.

\section{Related Work}

There are a number of works looking at automatic lay summarization in the biomedical domain \citet{goldsack_making_2022,guo_cells_2022,luo_readability_2022,devaraj_paragraph-level_2021}. One central issue is understanding the effect of text simplification on various aspects of summary quality such as relevancy. \citep{devaraj_evaluating_2022} explores the effect of text simplification on factuality by introducing a taxonomy of different error types that allows them to observe that while factual errors of missing information are a common error across generated and gold summaries, errors of substitution such as mixing up entities is a common occurrence in text simplification summarization models.

Supplementing source documents with external knowledge has been one of the main interventions discussed for mitigating factual errors and hallucinations in natural language generation systems \citep{shuster_retrieval_2021, lewis_retrieval-augmented_2021}.  With the idea that having access to grounding sources allows models to draw on those sources when generating text rather than being forced to rely on parametric knowledge which may be flawed \citep{thoppilan2022lamda,mallen2022trust}. In summarization, researchers have used factual knowledge from external sources to improve factuality by encoding external knowledge in models during training \citep{zhu_enhancing_2021, mao_fact-driven_2022} or to correct already generated summaries \citep{lee_factual_2022}. 

\citet{guo_cells_2022} evaluates retrieval augmentation as a method for enhancing abstracts with background knowledge. They use definitions from UMLS and Wikipedia as different retrieval corpora and find their method improves both the readability and relevancy of summaries while maintaining a similar level of factuality as models not using grounding. However, they did not perform retrieval-augmented generation for the lay summary generation task which is the novel contribution in this work.

\section*{Limitations}

Retrieval augmentation adds complexity to natural language generation requiring a separate retrieval module before the text generation step can begin. Additionally, retrieval augmentation possibly introduces more input text than the original input which is problematic for many neural network architectures with limited input space especially in the case of summarizing entire scientific papers. Finally, retrieval augmentation itself could introduce factual or relevancy errors if the retrieved documents are irrelevant or incorrect and they end up being used in the generated summary.

\begin{table*}[ht!]
\centering
\begin{adjustbox}{center,max width=\linewidth}
\begin{tabular}{@{}lllllllll@{}}
\toprule
            & DCRS         & FKGL         & bartscore    & summac      & bertscore    & rouge1       & rouge2       & rougeL       \\ \midrule
LED 8k        & 12.36 &  15.51 & \textbf{-2.21} & \textbf{21.37} &  86.22 & 45.01 & 15.44 & 24.15 \\
LED 16k       & \textbf{12.24} &  \textbf{15.32} & -2.22 & 21.34 &  \textbf{86.26} & \textbf{45.35} & \textbf{15.58} & \textbf{24.35} \\
All           & 12.29 &  15.42 & -2.28 & 21.36 &  \textbf{86.26} & 45.31 & 15.75 & 24.47 \\  \bottomrule
\end{tabular}
\end{adjustbox}
\caption{\label{tab:grounding_comparison_2}
Readability, factuality, and relevancy scores for our methods measured against a LED baseline trained with 8k and 16k tokens of the paper to be summarized respectively.  Lower is better for DCRS and FKGL and higher is better for the other scores.
}
\end{table*}

\bibliography{anthology,custom,Bibliography}
\bibliographystyle{acl_natbib}

\appendix

\section{BioLaySum}
\label{app:biolaysum}

The BioLaySum \citep{goldsack_making_2022} dataset has two splits, one for articles and lay summaries from PLOS and the other for articles and summaries from eLife. There are 29,119 training samples (24,773 from PLOS and 4,346 from eLife) and 1,617 validation samples (1,376 from PLOS and 241 from eLife). There is an additional test set but this was not used as the reference summaries are not publically available. All results are presented using the validation set. Our zero-shot summarization experiment uses a random sample of 300 summaries from the PLOS validation sample set.

\section{Model Training and Inference}
\label{app:models}
In our experiments, we use the Longformer Encoder-Decoder (LED) \citep{beltagy_longformer_2020}. We use the \texttt{allenai/led-base-16384} checkpoint on huggingface. All models are trained using 16,384 input tokens and an output length of 512 tokens except for \textit{LED 8K} which is trained using 8,192 input tokens. All models are trained on 4 v100 GPUs. Models are trained for 4 epochs and evaluated every 5,000 steps. The checkpoint with the best Rouge2 is selected as our best model. We set global attention to the first token and on a \texttt{<|SEARCH|>} token the model if the source document was supplemented with retrieval results. For inference, we sample text from the model using the following parameters with greedy decoding: number of beams 4, min length 100, length penalty 2.0, early stopping True, no repeat ngram size 3. We present LED 8k as the baseline comparison model in the main paper so that all models use the same amount of original article tokens. For a picture of what LED trained with 16k tokens from the original article see LED 16k in table \ref{tab:grounding_comparison_2}.

The crossencoder we used for re-ranking passages according to semantic similarity with the first 1,024 tokens of the original document was \texttt{cross-encoder/ms-marco-MiniLM-L-6-v2} using the sentence transformers library \citep{reimers_sentence-bert_2019}.

\section{Grounding Sources}
\label{app:grounding_sources}

\begin{table}[]
\centering
\begin{adjustbox}{center,max width=\linewidth}
\begin{tabular}{@{}lllllllll@{}}
\toprule
Source & Mean & STD \\ \midrule
Abstracts   &      8.81 & 3.0 \\
Wikipedia &    6.66 & 2.4 \\ 
UMLS &          2.56 & 2.1 \\
Wiki Simple &    0.71  & 0.8 \\
\end{tabular}
\end{adjustbox}
\caption{\label{tab:grounding_distribution}
    Distribution of grounding sources selected by our method (section \ref{sec:method}). Scientific Abstracts and Wikipedia are selected far more often than other sources.
}
\end{table}

APA reference strings were constructed using Crossref\footnote{https://www.crossref.org} using the first author name, title, and year from Crossref metadata matched on the BioLaySum document IDs. The UMLS definitions were from the sciscapcy entity linker \citep{neumann_scispacy_2019}. The Wikipedia index was a prebuilt index \texttt{enwiki-paragraphs} from Pyserini. We constructed the Wikipedia Simple index using the march 1st dump \footnote{https://dumps.wikimedia.org/simplewiki/20230301/}. The index was constructed extracting the plain text from each article and chunking articles into 6 sentence chunks. The 6 sentence chunks were indexed with Pyserini \citep{Lin_etal_SIGIR2021_Pyserini}. The Scientific Abstracts index was an Elasticsearch index of Crossref abstracts, it was accessed through scite.ai search api \footnote{https://api.scite.ai/docs}. Table \ref{tab:grounding_distribution} illustrates the distribution of the number of passages that have been selected for the grounding context of the \textit{Grounded} model in table \ref{tab:grounding_comparison}.

\section{Zeroshot summarization}
\label{app:gpt}

For our zero-shot summarization setting we use \texttt{GPT-3.5-turbo} with a temperature of 0. We evaluate on the entire BioLaySum so we can compare with results in table \ref{tab:grounding_comparison}. We prompt the model with the following prompts:

\noindent\fbox{%
    \parbox{\linewidth}{%
        \small
        \textbf{\underline{System Prompt:}}\\ You are a document summarizing agent focusing on summarizing documents to make them readable for a lay audience. Summarize the documents presented by the user in as simple terms as possible.\\
        \textbf{\underline{Prompt:}}\\ Summarize this document for a lay audience:
        \\
        \{document\}
        \\
        Below are a set of search results that ground the above document.
        \\
        \{search results\}
    }%
}
\section{Sample Summaries}

The below are samples of generated summaries of the article with the ID journal.pgen.1002882 from the PLOS validation set. \textit{Original} indicates the original summary.
\clearpage
\noindent\fbox{%
    \parbox{\textwidth}{%
        \small
        \textbf{\underline{Original:}}\\ 
        Messenger RNAs carry the instructions necessary to synthesize proteins that do work for the cell . Extending beyond the protein-coding sequence of a given mRNA is an additional stretch of sequence , harboring signals that govern how much protein is made and how long the mRNA remains in the cell before it is broken down . The incorporation of this end region into mature mRNA is itself subject to change; for the vast majority of human genes , how and why cells use different mRNA ends remains largely unknown . In this work , we surveyed mRNA ends from 10 , 000 genes in immune cells from genetically distinct human individuals . We found that mRNA end positions were not randomly distributed , but rather preferentially flanked the locations of regulatory signals that govern mRNA fate . The usage of these mRNA length forms and regulatory elements varied across individuals and could be dissected molecularly . Our results uncover key mechanisms and regulatory effects of transcript end processing , particularly as these are perturbed by genetic differences between humans .
    }%
}

\noindent\fbox{%
    \parbox{\textwidth}{%
        \small
        \textbf{\underline{Grounded:}}\\ 
        Gene expression varies widely between individuals of a population, and regulatory change can underlie phenotypes of evolutionary and biomedical relevance. A key question in the field is how DNA sequence variants impact gene expression, with most mechanistic studies to date focused on the effects of genetic change on regulatory regions upstream of protein-coding sequence. By contrast, the role of RNA 3-end processing in regulatory variation remains largely unknown, owing in part to the challenge of identifying functional elements in 3 untranslated regions. In this work, we conducted a genomic survey of transcript ends in lymphoblastoid cells derived from genetically distinct human individuals. Our analysis mapped the cis-regulatory architecture of 3 gene ends, finding that transcript end positions did not fall randomly in untranslation regions, but rather preferentially flanked the locations of regulatory elements, including miRNA sites. The usage of these transcript length forms and motifs varied across human individuals, and polymorphisms in polyadenylation signals and other 3 motifs were significant predictors of expression levels of the genes in which they lay. Our findings suggest that the strategy of integrating observed 3 transcript ends with inferred 3 regulatory motifs will prove to be a critical tool in continued efforts to interpret human genome variation.
    }%
}

\noindent\fbox{%
    \parbox{\textwidth}{%
        \small
        \textbf{\underline{LED:}}\\ 
        Gene expression varies widely between individuals of a population, and regulatory change can underlie phenotypes of evolutionary and biomedical interest. A key question in the field is how DNA sequence variants impact gene expression, with most mechanistic studies to date focused on the effects of genetic change on regulatory regions upstream of protein-coding sequence. By contrast, the role of RNA 3-end processing in regulatory variation remains largely unknown, owing in part to the challenge of identifying functional elements in 3 untranslated regions. In this work, we conducted a genomic survey of transcript ends in lymphoblastoid cells from genetically distinct human individuals. Our analysis mapped the cis-regulatory architecture of 3 gene ends, finding that transcript end positions did not fall randomly in untranstranslated regions, but rather preferentially flanked the locations of regulatory elements, including miRNA sites. The usage of these transcript length forms and motifs varied across human individuals, and polymorphisms in polyadenylation signals and other 3 motifs were significant predictors of expression levels of the genes in which they lay. Our results underscore the importance of 2 mechanisms at play in the genetics of gene expression: the usage of distinct 3 end processing signals and the effects on 3 sequence elements that determine transcript fate. Our findings suggest that the strategy of integrating observed transcript ends with inferred 3 regulatory motifs will prove to be a key component in the systems-level modeling of regulatory networks and their variation.
    }%
}

\noindent\fbox{%
    \parbox{\textwidth}{%
        \small
        \textbf{\underline{Zero-shot UMLS:}}\\ 
        This document talks about how genetic differences between individuals can affect how genes are expressed, which can have important implications for evolution and medicine. Specifically, the document focuses on how differences in the way RNA molecules are processed at their ends can impact gene expression. The researchers used a technique called 3'-end RNA-seq to study this process in human cells, and found that variations in the processing of RNA ends can be influenced by genetic differences between individuals. The study sheds light on the regulatory elements that control gene expression and suggests that understanding these elements will be important for interpreting human genome variation.
    }%
}

\end{document}